\documentstyle[harvard,a4,11pt]{article}
\title{\bf Automatic summarising: factors and directions}
\author{
Karen Sparck Jones\\
Computer Laboratory, University of Cambridge\\
New Museums Site, Pembroke Street\\
Cambridge CB2 3QG, England\\
sparckjones@cl.cam.ac.uk}
\date {May 1998}

\begin{document}

\maketitle

\begin{center}
To appear in\\
{\em Advances in automatic text summarisation},\\
Ed. I. Mani and M. Maybury, Cambridge MA: MIT Press, 1998.
\end{center}
\vspace{10mm}

\begin{abstract}

   This position paper suggests that progress with automatic summarising
demands a better research methodology and a carefully focussed research
strategy. In order to develop effective procedures it is necessary to
identify and respond to the context factors, i.e. input, purpose, and
output factors, that bear on summarising and its evaluation. The paper
analyses and illustrates these factors and their implications for
evaluation. It then argues that this analysis, together with the state of
the art and the intrinsic difficulty of summarising, imply a nearer-term
strategy concentrating on shallow, but not surface, text analysis and
on indicative summarising. This is illustrated with current work, from
which a potentially productive research programme can be developed.

\end{abstract}

\section{Introduction}

   This paper addresses two pressing questions about automatic summarising:
given the present state of the art, what are the best directions to take in
research methodology, and in strategy development? The paper reviews where
we are now in automatic summarising, and argues that our methodology focus
should be on {\em context factors}, our strategy focus on {\em shallow
processing}. My claims are that we will not be able to develop useful
summarising systems unless we pay proper attention to the context factors,
and especially purpose factors, that shape summaries; and that since we
cannot expect even in the medium to long term to emulate human
summarising, we should concentrate, in seeking generally applicable
procedures, on relatively shallow techniques that are within reach of
current NLP technology. My two concerns are related because the limitations
of the technology imply a need for careful identification, from a context
point of view, of the summary tasks to which even a quite limited technology
could add value, and of the conditions under which it could be successfully
applied.

   The paper is a programmatic one. My report on the present state of the
art is therefore designed only to note its salient features, and does not
review recent and current work in detail; and my argument for where we 
should
go is intended as a call to arms, and is therefore broadly assertive, with
indicative illustrations, but not finely elaborated.

\section{Background}

   As background, I shall take for granted an initial definition of a
{\em summary} as

\noindent
{\em a reductive transformation of source text to summary text through
content reduction by selection and/or generalisation on what is important
in the source}.

\noindent
I shall also assume that the basic process model is a three-stage one:

\begin{tabbing}
ZZZZ\=ZZZ\= \kill
\>I \>:\ \ source text {\em interpretation} to source text representation\\

\>T \>:\ \ source representation {\em transformation} to summary text
representation\\

\>G \>:\ \ summary text {\em generation} from summary representation.\\
\end{tabbing}

   This model (slightly modifying the earlier one presented in
Sparck Jones 1995)
may appear obvious. But adopting a general framework with distinct
processing phases and data types supplies a useful common means for
checking the real logic underlying specific systems, making it easier to
identify the assumptions on which they are based, and to compare
one system with another. Of course each major stage can subsume several
substages, for instance in interpretation a process for building
individual sentence representations, followed by one integrating these
into a larger text representation, perhaps followed by a further process
modifying the global text representation. The nature of the assumptions
on which a summarising system are based are made clear when it is
recognised, as for example with DeJong's FRUMP (DeJong 1982), that the 
source
and summary representations are conflated.

   The definition of a summary, though no more than an informal and
obvious one, nevertheless serves to emphasise the fact that summarising
is in general
a hard task, even if we can find, and expect to automate for,
some situations
where it is not. Summarising is hard because we have to characterise a
source text as a whole, we have to capture its important content,
where content is a matter both of information and its expression, and
importance is a matter of what is essential as well as what is salient.

\section{Current state}

  Research on automatic summarising, taken as including extracting,
abstracting, etc., has a long history, with an early burst of effort in
the sixties following Luhn's pioneering work, two subsequent decades
with little research and a low profile, and a marked growth of activity
since the mid eighties and especially very recently:
see Paice (1990), Endres-Niggemeyer et al. (1995),
IPM (1995), Mani and Maybury (1997). But virtually
all of the work done so far, and especially the more
practically-oriented work involving substantial rather than suggestive
implementation, falls under two headings: {\em text extraction}
and {\em fact extraction}.

   In text extraction, where `what you see is what you get', some of
what is on view in the source text is transferred to constitute the
summary text. Text extraction is an {\em open} approach to summarising,
since there is no prior presumption about what sort of content
information is of importance. What is important for a source text is
taken as marked by the source text according to the general,
linguistically-based,
importance criteria applied in the extraction process. With fact
extraction the reverse is the case: `what you know is what you get',
that is, what you have already decided is the sort of subject content
to look for in source documents is what you seek to extract. It is a
{\em closed} approach in that the source text does no more than
provide some instantiation for previously-established generic
content requirements. The text extraction method is intended to let
important content emerge, as individually appropriate from each source.
The fact extraction method is intended to find individual manifestations
of specified important notions, regardless of source status.

   (Much recent work has been on generating text summaries from
non-linguistic source data
(e.g. McKeown et al. 1995). This is practically an important variant
of summarising, and may be far from easy, but I am excluding it here
because the critical issue for summarising in general is the way the
(long) source text is interpreted so that important content can
be recognised and `lifted' from it.)

   The processing techniques that characterise the two
extraction approaches are
naturally very different. In text extraction, processing effectively
merges the interpretation and transformation stages. Key text segments
(usually whole sentences) are identified by some mix of statistical,
locational, and cue word criteria; and generation is primarily a matter
of smoothing, for example by incorporating source sentences preceding
key ones containing anaphoric references. One possible way of viewing
this type of strategy,
as it is often implemented,
is to say that the source text is taken as its
own representation, without any interpretation, and this representation
is then subject to a transformation stage which is simply extractive.
The output summary is therefore close to the source in linguistic
expression, and also structure as far as presentation order is concerned.
In general, with summaries produced with this strategy, the source
is viewed through a glass darkly. The selected sentences together
usually have some relationship to what would independently be judged as
important source content, and allow the reader to infer what it might
be. But this dim view of the original is typically also made more
obscure because the output summary text, even after smoothing, is
itself not very coherent.
Even where an explicit derived or abstract representation, e.g. a
frequency-annotated word list, is formed and used for sentence
extraction, these points about the resulting summary apply.

   With fact extraction (and its variants, for example ``message
understanding'') the interpretation and transformation stages
are also essentially merged. The initial text processing is designed
only to locate and process the source text expressions that bear on
the specified generic concepts, or concept relations, for which
particular factual instantiations are sought. There is no independent
source text representation, only the direct insertion of source material,
with more or less modification of its original expression according to
the individual application requirements, in some information
representation that is usually of a frame (template, schema) type.
Transformation is already all or largely done, via the initial selection
of source material and frame filling. In particular, while individual
source items may be substantially transformed for frame entry, the
invocation of a predesigned frame constitutes the transformation of
the source as a whole. Generation on the other hand usually involves
the proper production of natural language text from the underlying
representation. The essential character of this approach is that
it allows only one view of what is important in a source, through
glasses of a particular aperture or colour, regardless of whether
this is a view showing what the original author would regard as
significant.

   In practice, there is considerable variation in each of these
approaches, often associated with the degree of source reduction.
Thus for short sources, single key sentence extraction may be
reasonable (if risky), and avoids the problem of output coherence.
Similarly, with types of short source, it may be appropriate to
process virtually all of their factual content, as in Young and
Hayes (1985)'s work on banking telexes. On the other hand where summary
generation within the fact extraction  paradigm is from multiple input
sources, there may be more transformation of their combined
representation, as with POETIC
(Evans et al. 1996), where summarising is dynamically
context-dependent.

   But overall, the two styles of summarising that dominate
current research are complementary. The text extraction approach has
the advantage of generality, but delivers low-quality output
because the weak and indirect methods used are not very effective
in identifying important material and in presenting it as well-organised
text. The fact extraction approach can deliver better quality output,
in substance and presentation, as far as the selected material is
concerned, but with the disadvantages that the required type of
information has to be explicitly (and often effortfully) specified
and may not be important for the source itself.

   Clearly, it is necessary to get more power in automatic summarising
than text extraction delivers, and more flexibility than fact extraction
offers. This applies even if some customisation effort is acceptable,
as is required for the fact approach, or might assist the text one.

   Taking the larger view, it is evident that as summarising is all
about the reduction of extended source text, it is necessary to address
the role of large-scale discourse structure in signalling important
text content (as well as its role in supporting discourse interpretation
in its own right). General advances in automatic summarising will
therefore require methods of capturing this structure in source
interpretation, and of deploying it to support the condensing
transformation stage, as well as making it available for use in
generation. This need would imply, for example, going further in
exploiting Rhetorical Structure Theory than Marcu (1997)'s RST-based text
selection strategy does, since we have to envisage using
large-scale discourse structure to transform source content.
The same applies to other approaches to capturing discourse
structure, like those presented in Hearst (1994) and Hahn and Strube
(1997).

   But though understanding how to identify and use large-scale
discourse structure is critical for serious summarising that can deliver
summaries of the same quality as human ones, it is a long-term research
topic. It involves, for example, exploring alternative views of
large-scale discourse structure where, because theories of discourse
structure are not well developed, there are many questions to answer even
before the attempt is made to apply structural information in summarising
(Endres-Niggemeyer et al. 1995, Sparck Jones 1993).
These questions are about: the types of information - linguistic,
attentional, communicative, domain - that defines structure; the forms
of structure - instantiated, constructed; the indicators of these sorts
of structure, their functional roles, their relationships and, of
course, their identification and use. Finding
their answers will manifestly
take time, and
it is therefore rational to look, in parallel, for ways of
achieving nearer-term progress in automatic summarising.

\section{Methodology}

   I believe that in order to make progress in devising effective
{\em general} summarising strategies, even if only modest ones, we
need to pay much more attention to {\em context factors}, i.e. to
identifying the operational factors for any particular application.
It is possible, in building a summarising system for some application,
simply to try it and see - taking whatever strategy we have to hand
and discovering whether it delivers outputs that users can accept; and
this seems to be what is often done in practice. But it is a crude method
of working, and determining the operational factors for individual
cases should encourage both a more rational choice of strategy and
better understood strategy development.

   It is important to recognise the role of context factors because the
idea of a general-purpose summary is manifestly an {\em ignis fatuus}.
When the range of summarising contexts is considered, there is no
reason to suppose that any one summary, even a supposedly
good one, would meet all the context constraints, even if only
moderately well. Why should any one specific technique give the
best or even an acceptable
result regardless of the properties of the input source or the
requirements for the output summary? Similarly, the notion of a
{\em basic} summary, i.e. one reflective of the source, makes hidden
factor assumptions, for example that the subject knowledge of the 
output's readers will be on a par with that of the readers for whom the
source was intended.
This is natural for abstracts prefacing papers, but does not always
apply, and either way, source properties need identification so their
implications for summarising can be handled.
Effective summarising requires an explicit, and
detailed, analysis of context factors, as is apparent when we recognise
that what summaries should be like is defined by what they are wanted for,
as well as by what their sources are like.

\section{Context factors}

   It is convenient to distinguish three classes of context factor:
{\em input}, {\em purpose} and {\em output} factors. There is a major
problem in that all the factors and their various manifestations are
hard to define, so capturing them precisely enough to guide summarising
in particular cases, or to provide the foundations to guide strategy
development, is very hard. The factor presentation that follows is
therefore primarily intended to emphasise the range and richness of the
influences on, and hence varieties of, summarising. I shall first list
the factors and their possible realisations, with brief examples, and
then give a larger illustration for a single application.

\subsection{Input factors}

  These fall into three classes: source {\em form}, {\em subject type}, and
{\em unit}.

  Input {\em form}
in turn subsumes source text {\em structure}, {\em scale},
{\em medium} and {\em genre}. Structure includes both
explicit organisation as marked by subheadings
(e.g. Objective, Data, Method ...), and also structure embedded in text
like familiar rhetorical
patterns (e.g. statement followed by elaboration).
It may or may not be appropriate to preserve this structure in the output
summary: thus not merely source headings but also the associated grouping
of material may be abandoned. Scale, given that we may talk of summarising
a paragraph or
a book, is important because it has implications not only for the degree
of reduction in summarising, but also for the possible extent of content
transformation.
Thus a long novel can be more variously viewed
and presented in summary than a short news story.
Medium covers both different natural languages
and sublanguages: it is not necessarily the case that the source
sublanguage should be preserved in the summary, as for example when
telegraphese is expanded to normal prose. Genre, an ill and variably
defined notion, here refers to literary form rather than content type, for
instance description or narrative. While characterisations like these
are very vague, we must expect genre to affect summarising, for example
with a narrative implying a decision
either to preserve the presentational
sequence or to exhibit the real event sequence, or
perhaps to select major
events while omitting linking ones.

   Thus a specific source text, e.g. a company project progress report,
may have a structure defined by task labels, modest scale, much
company-specific nomenclature, and descriptive accounts of present
task states. Questions for summarising could then be whether the given
task organisation be retained or replaced, say by one grouping tasks
as complete and incomplete; whether a very brief one-sentence summary can
be produced for such a moderate-length source without too much information
loss; whether department-level acronyms be spelt out; 
and whether the simple
state description genre of the source should give way to a checklist
rhetoric.

   Of course, there have to be reasons for answering these questions
one way rather than another: this is where the purpose factors considered
below play
their key part; and there may also be dependencies between the answers.
The point here is that there is no mandatory replication of source
text properties in summary text ones; but that decisions on the properties
that summaries should have, as these are fed by purpose requirements,
imply that the properties of the source are recognised. This applies whether
or not these properties are to be changed. While it is possible to adopt
an essentially passive summarising strategy, and perhaps to implement this
sufficiently effectively without explicitly identifying source text
properties, such a strategy carries with it (as noted earlier)
assumptions about the context in which the summary is to be used. To
meet stated summarising requirements it is necessary to identify source
properties explicitly, so that the summarising transformations
can be properly carried through. This applies in principle even though,
as noted earlier for current fact extraction techniques, summarising to
meet output requirements may ride roughshod over sources. Thus
{\em in general} we have to capture source text properties because
processing has to respond to them, even if they may in specific cases
vanish before the output text is reached; however
as we are summarising unique communicative texts, we would expect to
preserve their distinctive properties unless
countermanded; and doing this effectively will normally, to avoid
distorting transformations, imply that these properties are preserved
in the source representation.

   Source text {\em subject type} may be broadly characterised as
{\em ordinary}, {\em specialised}, or {\em restricted},
in relation to the presumed subject knowledge
of the source text readers. The distinction between specialised and
restricted is between subject matter that depends on knowing what
people in many different locations may be assumed to know and subject
matter that is limited to some specific local community. Source texts
giving chemical analyses as found in journal publications illustrate
the specialised case, ones that rely on local names and references
the restricted case.  Subject matter also of course covers
many individual domains with their own particular knowledge, e.g.
sport, gardening. The subject matter factor clearly bears on summarising
through the assumptions that are made about background subject
knowledge in summary users, as opposed to source readers. For instance
heavily technical financial information in a source may be radically
simplified for more popular consumption.

   The final input factor, {\em unit}, distinguishes summarising over a
{\em  single} input source from summarising over {\em multiple} sources.
More
particularly, this is a distinction between summarising where the
source, even if it is collection as in an edited book, consists
of material previously brought together with design as a whole and
summarising where the sources are independently generated and not brought
together with consideration, as for example a succession of press
wires on some major event. Though, as with all the previous
distinctions, this is not a hard and fast one, the implications for
summarising of single versus multiple sources are in the treatment of
information redundancy or of changing data over time. In the latter
case it may be further necessary to consider, in the light of summary
purpose requirements, whether only the latest data be treated, or the
changes themselves indicated.

\subsection{Purpose factors}

   These are the most important factors. It may seem obvious that they
should be explicitly recognised and their definition applied to guide
the choice of summarising strategy. But in practice in automatic
summarising work they are often not recognised and stated. This may be
because it is tacitly assumed that the purpose of summarising and its
implications are obvious, whether this is because 
summarising is viewed as
requiring `straight' content condensation as an operation
regardless of context, or
is taken to be aimed at basic reflective summarising, or
because some processing method or form of
output is specified which is already assumed as appropriate for
the summary purpose. But since the purpose factors are critical, their
implications may be ramified, and as they are the basis for evaluation,
it is essential to analyse them properly.

   Purpose factors fall under three headings: {\em situation},
{\em audience}, and {\em use}.

   {\em Situation} refers to the context within which the summary is to be
used, which may be labelled {\em tied} or {\em floating}.
The former
refers to cases where the particular environment within which the
summaries are to be used - who  by, what for, and when - is known
in advance so that summarising can be tailored in a detailed way
to meet these context requirements: for example, product
description summaries adapted for the use of a company's marketing
department for a particular sales drive. A floating situation is
where there is no precise context specification. Thus technical
abstract journals may be quite narrow in their view of other purpose
factors, but nevertheless not be tied to predictable
contexts of use.

  The second summary factor, {\em audience}, refers to the class of reader
for whom summaries are intended. Audiences may be more or less
tightly characterised along a spectrum from
{\em untargetted} to {\em targetted}
in terms of assumed domain knowledge, language skill, etc. Thus
the readers of a mass market women's magazine may be deemed untargetted
with respect to summarising a fiction serial, since they will be
so varied in experience and interests. The audience for summaries in
a professional abstract journal, say in the law, on the other hand, is
(at least by comparison) targetted.
There are,
further, many possible specific audiences with different characteristic
interests, for instance implying different summaries of the same
manufactured product report. Defining the summary audience is important
because,
as already noted, it should not be taken as similar to the audience for
the source.

   The third purpose factor is {\em use}: what if the summary for? Possible
uses for summaries include those as aids for {\em retrieving} source text,
as means of
{\em  previewing} a text about to be read, as information-covering
{\em substitutes} for their source text, as devices for
{\em refreshing} the
memory of an already-read source, as action {\em prompts} to read their
sources. For example a lecture course synopsis may be designed
for previewing the course, and may therefore emphasise some information
e.g. course objectives, over others. 

   The account of purpose factors just given is, like that of input
factors, only a beginning. It is clearly necessary to develop
a fuller view and characterisation of purpose factors, as a basis for
building generic systems with the capacity to meet purpose requirements
and for analysing individual applications in  order to determine
their requirements; and it is then necessary, for an application,
to do this analysis. This can be expected to be much more detailed
than the examples above, and to have quite detailed implications for
the treatment of source text content in summarising. Thus if we
know that the summarising audience is a particular scientific community,
with a particular type of use for summaries in a particular type of
situation, this could imply that one specific kind of scientific
claim needs to be identified as the major claim in a source paper.
As this also suggests, summarising may not only rely on background
domain knowledge and even import this into a summary where it is
presumed but not given in the source (as in some frame-based approaches):
it can rely on other forms of background knowledge and also import
this, e.g. stating the situation for which a summary is produced.

\subsection{Output factors}

   The final major class of factors is output factors. There are
at least three major ones, namely {\em material}, {\em format},
and {\em style}.

   The {\em material} factor refers to the extent to which the summary is
intended to capture all of the important content in the source text,
or only some aspects of it. The default case is that the summary is
a {\em covering} one (subject to some concept grain level). But
summaries may be designed to capture only some types of source
information, for instance in astronomy papers what was observed,
or for fiction only the plot, as well as only some concept instantiations
as considered under fact extraction. These summaries are intentionally
{\em partial}.

   For the second output factor, {\em format},
we can see a broad distinction
between summaries that are wholly or primarily {\em running} text,
as in many journal paper abstracts, and those that are {\em headed}, where
the summary material is tagged or organised by fields that may
indeed be standardised across summaries, for example using a `Test results'
heading in biological abstracts.

   The third output factor is {\em style}. For example a summary may be
{\em informative}, conveying what the source text says about something;
{\em indicative}, noting that the source is about some topic but not
giving what it says about it; {\em critical}, as in a summary that
reviews the merits of the source document; or {\em aggregative}, used here
to define summaries where varied sources, including multiple ones
of the same type,
are deliberately set out in relation to one another, as in a judicial
summing up. This is not an exhaustive 
(or exclusive) list of alternatives, only a start.

   These
output factors
have also just been presented as free properties of an output
summary text. But they of course follow from judgements about what
the summary should be like given the nature of the input source and the
purpose the summary is intended to satisfy. Thus a summary may only
partially cover a source text because the purpose of the summary is
to supply a certain kind of information, in condensed form, for a
certain kind of use, regardless of other elements in the source.
The relations between the three factor types can therefore be
expressed as defining a summary {\em function}, which is:
given Input Factor data, to satisfy Purpose Factor requirements,
via Output Factor devices.

   As this suggests, a quite full ground analysis of input and purpose
factors is needed to reach the desired output properties for
summaries. But at the same time the unavoidable indeterminacy of
summary use (because summaries are made for {\em future} use) implies
that the joint input and purpose characterisation cannot simply mandate
a specific choice of output factor properties. There will be a need
for judgement as to the likely best ways of meeting the purpose
requirements given the data characteristics, though if the input
and purpose analysis has been well done the choices for output will
be limited. These are in any case generic choices, they have to be
carried through for each individual summary, applying primarily to
the summarising transformation and following through
for the output generation.

\section{An illustration}

\hspace{5mm}
{\em Book review summaries for librarian purchasers (in public libraries).}

   We assume a library service which distributes information
about new books. This consists not only of basic
information like bibliographic details and general type, for
instance biography, history, fiction, but also summaries of reviews
published in e.g. literary weeklies. The aim is to help librarians
decide what to buy for their libraries.

   The input factor characterisation for this application is that
the source texts (i.e. original reviews) have a form that is
essentially simple running text; are variable in scale; have literary
prose as their medium; and are single units. The purpose factors
are a floating situation, since the summaries are distributed
on a library mailing list without knowledge of their individual readers,
or the precise circumstances in which they are used (e.g. what other
information is combined with the mailing to determine purchasing); an
untargetted audience, since public librarians vary widely though a
general and professional education can be assumed; and summary use as
a substitute for the original review, to which the librarians may not
have ready access. Rational choices for the output summaries that
follow from this input and purpose analysis are that the summaries
should be covering ones, not selecting only some types of information
from the original reviews; be delivered as simple running text attached to
the bibliographic header supplemented by a specification of the source
review location and writer; and that the style should be indicative
of what aspects of the book the review considered and what attitude
the reviewer took to it.

   Given this starting position, but only from such a starting position,
the details of how to produce the required summaries can be addressed.

\section{Evaluation}

   A context factor analysis like that just listed is needed for
appropriate strategy selection and parametrisation. It is also crucial
for evaluation. It is impossible to evaluate summaries properly without
knowing what they are for.

   The two conventional approaches to evaluation, while not without
some utility, are much less useful than might be supposed. One is
to evaluate summary texts by comparison with their source texts, in an
attempt to answer the question: Has the summary captured the important
concepts (and concept relations) in the source? The problem with this
approach is how to tell whether it has or not. Even if humans are
rather better than current machines at identifying important concepts
in discourse, it does not follow that they can be laid out or checked
against one another in any simple way. or indeed that this human
analysis will not introduce its own biases or have its own defects.
The expression of the key concepts for comparative purposes gives
language objects that are subject to all the variable interpretation
that such
objectives have, while their relation to their respective source and
summary texts is inaccessible and is precisely what the summarising
process is seeking to capture. Manuals for professional abstracters
(e.g. Rowley 1982) recommend that abstracters go over the source to check
they have got the main points from it, but this assumes the capability
we are trying to define.

   In any case, as the discussion of factors brought out, the relation
between source and summary need not be `just' a reduced replication
of the same key concepts. As the discussion implies, my original
definition of summarising has to be extended to allow for the use of
and presentation in the output summary of information not in the original,
and of the introduction of new perspectives on the original. More
importantly, my definition of summarising has to be extended,
as a general definition, to
recognise the role of summary purpose in determining the nature of
the content condensation.

   Thus while this comparative method of evaluation,
setting summary against source,  can be
helpful as a development, or rapid vetting, tool, it does not provide
an adequate base for rigorous evaluation.
The same applies to the more limited, albeit more controllable,
question method, designed to establish whether questions (about
key content) that can be
answered from the source text can also be answered from the summary.

   The main alternative evaluation strategy is to compare automatically
produced summaries with human ones for the same source. But again,
even if we assume that the conditions of summarising are understood
and are the same in both cases, when dealing with content rather than
extracted text determining that two texts share the same key ideas
is far from trivial. This strategy moreover assumes that the human
summary is itself the best reference standard, which is not necessarily
the case, as many analogous studies of document indexing have shown.
Thus automatic index descriptions of quite different types may
be as effective for retrieval as manual ones (see e.g. Salton 1972,
1986; Cleverdon 1977). This `human reference' approach to evaluation
should therefore, like the previous one, be treated only
as a rough, preliminary evaluation tool.

   In either case, moreover, to make proper use even of these tools, it
is essential to take into account what the summary conditions are.
Comparison between source and summary for key concept capture
should be modulo the pertinent context factors, and the same for
comparisons between human and automatic summaries. It is evident,
therefore, that it is much better to adopt an evaluation strategy that
refers directly to the context constraints and in particular is based
on summary purpose. This applies even where it may be thought legitimate
to see whether one summary is like another, because the latter has already
been shown suited to purpose: there is enough latitude in summarising
for it to be better to adopt the direct test for purpose.

    This still, however, leaves much to be done not only in specifying
the application constraints on summarising in an individual case,
but in the design and execution of the evaluation, notably the
precise definition
of evaluation goals and methodology including measures, data, procedures
and so forth
(Sparck Jones and Galliers 1996).
For instance with the library purchasing example, evaluation
could be from quite different perspectives, and a quite different
levels of detail. Thus it could progress from
very simple but still useful starting points, for instance answering
just the question: Does the output summary include all
the administrative information it ought, which can be rather easily
checked. Then, making reference to the purpose for which the summaries
are intended in a more direct and comprehensive way, an evaluation could
be designed to answer the question: Does the summary allow librarians
to reach decisions about whether to buy the book in question or not
(though this strategy would have to allow for the fact that it could be the
original review, and not the summary, that is defective if there is
no decision.) Such an evaluation could be done by decision recording, say.
It in principle addresses the key issue as to whether the summary
transmits the kind of information pertinent to purchasing that could
be drawn from the source review, where the first evaluation addresses only
the system's ability to transmit supporting administrative data. But
the key issue is still addressed only in rather a loose way. It would
seem that summary quality (i.e. utility) could properly only be established
by whether the librarians' decisions were `correct', i.e. that decisions
based on summaries were the same as those based on their source reviews.
But while this might be determined by decision recording, as before,
it would require careful sampling design to gather independent decisions,
and could require quite a large scale study to compensate for other
variables including, for instance, interaction effects with prices.

   These are only some illustrative possibilities. The point is that
it is always necessary to motivate whatever evaluation is conducted
by reference to the ulterior purpose for which the summaries are
intended,
whatever particular perspective is adopted for the evaluation,
and to work it out in detail. This is clearly a challenge for
cases where summaries are envisaged as for multiple uses, and varied
audiences, perhaps far in the future. The context factor analysis
is nevertheless always required, as it is vital to the design and
choice of summarising strategies, even if the devil is in the detail
of what follows from this analysis for strategy specification.

   But context analysis also has a critical role to play in helping us,
with the present state of the art in automatic summarising to work
from, to choose
research directions for summarising that would be both useful and
feasible.

\section{Research strategy}

   What, then, are the implications from the discussion of context
factors for an advance on more powerful, general summarising technology,
and specifically for
effective NLP techniques and system designs? In particular,
while it is overall appropriate to exploit the factor analysis as
a basis for developing much better methods of summarising than we
have now, discovering how to apply the analysis as a lever for
source text {\em discourse} interpretation is a major, long-term,
research enterprise. There is therefore every reason to pursue a
parallel research strategy of a more practically-motivated and realistic
sort aimed at developing useful summarising methods, for at least
some types of environment, in the short to medium term. This is subject
to the important requirement that these should be general methods, i.e.
not require heavy hand customisation as exemplified by approaches
relying on domain frames. Clearly, given this requirement, there will
be types of, or individual, applications for which we cannot expect
to supply a system; and we should allow, even with general strategies, for
some parametrisation. But we should try to give substance to the idea
of systems that at most require a little tailoring, or can be largely
assembled from standard parts, rather than built from scratch as far
as the central transformation component is concerned.

    This implies focussing on environments where indicative, skeletal
summaries are useful. That is, on environments where the user has a
rather loose or coarsely defined task, has other aids to carrying out
the task so high-class summaries of the kind exemplified by scientific
journal abstracts are not required, and indeed for which summaries
are not
essential. Thus we should focus initially on environments
where summaries are helpful
but not vital. These should also be environments where the user is
knowledgeable about the `information scene', and where the user operates
interactively so they can check or enlarge on  their
interpretation of a summary by reference to the background context,
relate the summary to other sources of information, formulate and
carry out their task in a flexible way, and so forth.

   The goal is therefore to supply summaries that are `sufficient for
the day', for example where the role of summaries is to facilitate
browsing in an interactive information retrieval environment where
there are other search tools available, as well as access to the source
documents; or to provide notification about source documents where it
is not necessary to do more than give some brief lead into source
content, but it is desirable to do more substantial than supply an
unstructured list of keywords. Both browsing and alerting are
generic activities that can figure within many different encompassing
task environments; there are many applications where providing
users with word lists, or repeating brief document titles
(assuming they exist), is not enough
and where a short, readable text indicating the substance of a source
document is really valuable.

\section{Forward direction}

   Some work is already underway of this broad kind
(e.g. Boguraev and Kennedy 1997). So my
proposal that we should embark on it may appear, if not redundant,
too unadventurous. But it is not really so. What is required is far
from trivial, once the source {\em text} extraction strategy is abandoned,
i.e. steps are taken towards source interpretation; and it can provide
a good grounding for progress towards fuller and deeper interpretation.
At the same time, as already claimed, success would be very handy right
away. Boguraev and Kennedy's approach, and others reported in
Mani and Maybury (1997), are still text based; so though they
are generally more sophisticated than earlier systems, and may deliver
useful outputs, these are still very limited as content summaries.
It is clearly necessary to look for something more substantial.

   Now to be more specific about the form nearer-term research
strategies
should take, i.e. about the type of method worth investigating (given
current NLP capabilities). It has to be better than surface text
extraction; but equally, for a generally-applicable system, cannot
seek deep information like `fact extraction' does.

   I believe that the right direction to follow should start with
{\em intermediate} source processing, as exemplified by
sentence parsing to logical
form, with local anaphor resolution. This is taking source text
analysis as far as it can be taken on a linguistic basis, without a
reference domain model. But with this initial
processing we can still expect to get more manifest discourse entities and
relations between them than the source text in  itself supplies, and at
the same time maintain a more sympathetic account of the original source
than with the fact extraction method driven from prescribed selective
fact types.

   Processing in this way would, if carried no further, give an
extremely shallow source representation as a sequence of sentence
representations. But these can be linked by common elements to give
a more integrated representation. This is still shallow, and unavoidably
so, but has important advantages. It embodies some interpretation of the
source text
as a whole; and it is neutral, i.e. it is not biased towards any
particular view of the source's content or prior specification
of the form and content of the eventual summary. The general
characteristics of such source representations, even on the rather
strong assumption that current engines can deliver them reasonably
reliably, and
that the problem of competing sentence analyses can be finessed,
clearly places some limits on what can be done in summarising: thus
without world knowledge, full discourse entity identification is
impossible. But there is still large scope for different ways of
exploiting the representation for summarising purposes, and in particular
scope for more than in approaches where the source representation
is directly constructed as the summary one.

   Thus we can seek to derive a summary representation using more than
one type of data, notably statistical data about the frequency of
entities (rather than just words), and markedness information, as
represented by text location, discourse focus,
or cue expressions.
Markedness data can be associated with the representation and
its discourse entities because the source representation is not
too divorced from the original text
(though precisely how far markedness data
are retained during analysis depends
in practice on the detailed style of a logical representation).
However its value is enhanced by the fact that it is not tied to
surface text strings as such, but to their propositional referents.

   Then since a summary representation will be some derived logical
form structure, it can be taken as an input to an output text generator.

   The core argument on which this suggested sensible research strategy
is based is that as full source text analysis is impossible at present
or in the nearer term, but robust parsing is feasible now, this will
get enough logical form predications for summarising because extended
text is {\em redundant}, especially with respect to key information.
If text content is important, and it therefore mattered to the source
writer that the source reader
should get this information, it will be emphasised
and reiterated. Then from the summarising point of view, we can assume
that even if the source representation is incomplete, both because our
analysers are only linguistic and because they are in practice imperfect,
it will nevertheless capture most, if not all, of what is important in
the source text. This will apply whether the representation is
incomplete but so to speak evenly so, through systematic limitations
on analysis, or is incomplete and unevenly so, because of specific
analysis failures. In the first case, what is important in the source
should still be relatively more prominent; in the second, what is important
will be retained one way or another.

   Clearly, without full source interpretation, we may expect to miss
some material detail, for example the quantification an initial
purely linguistic parsing into logical form does not usually deliver or,
as mentioned earlier, full co-reference identification. The question
is whether enough of the key source content will be captured for a fairly
sound if not wholly accurate summary. In particular, will this approach
lead to a better, if still somewhat schematic, picture of the source than
the text extraction technique?

   The challenge for the proposed strategy is in filling in the
necessary detail: building the source logical form representations
that capture discourse entities, their relations, and their relative
status; and deriving the summary representation from this; also,
evaluating the resulting summary texts for applications.
Evaluation will not be easy: if the summaries
are intended to assist tolerant but rational users with rather loosely
defined tasks and varied tools for executing these, this can be expected to
make evaluation to determine the
(comparative) merits of (alternative) summaries more, not less,
difficult.

\section{An illustration}

   To flesh out the proposal, though without claiming that what follows
is the only or best individual approach, I shall take work by Richard
Tucker in progress at Cambridge. This illustrates the kind of shallow
approach advocated, being studied within an experimental rig designed
to test a range of detailed particular variations on the generic
strategy, with unrestricted news story texts as sources.

   In the input interpretation stage of processing, sentences are
parsed into quasi-logical forms using the SRI Cambridge Core Language
Engine (Alshawi 1992). This delivers predications with as much
disambiguation as can be achieved by using purely linguistic resources,
but without quantification since this requires domain context information.
The sentence representations are then decomposed into their simple, atomic
predications, and these are linked to build up the overall
source representation as a {\em cohesion graph}. The graph nodes are
the simple predications, and links between them are of different types
based on common predicates, common arguments within sentences, or
similar arguments across sentences. The last is a relatively weak form
of connection, based on having the same semantic head: establishing
stronger connections presupposes the ability to determine actual
entity relations, which as already noted cannot
in general be achieved using only
linguistic means and without quantification structure.
The weaker,
primarily semantic relationships embodied in the
cohesion graph may nevertheless be adequate
for the level of summarising being sought:  even if the
specific discourse entities cannot be identified, the sort of
entity that is involved can, and this may be convey sufficient
information for summary utility: thus it may be enough to know
a source is about films, without knowing whether it is about
one or several. There may be a single
graph covering the whole source text, or perhaps several graphs.
The main desideratum is at least a fair degree of cross-sentence
linkage, and if the original is coherent
(if only because redundant), this can be expected.

   One advantage of the decomposition to atomic predications is that
even if there are alternative parsings for whole sentences, they
may share these constituents, so limiting representation to just
one, possibly incorrect, sentence parsing may not be too damaging.

   The transformation step, deriving the summary representation from
the source one, identifies the node set, i.e. predication set, forming
the summary content. It exploits weights for the graph edge types,
applying a scoring function seeking {\em centrality},
{\em representativeness},
and {\em coherence} in the node set, i.e. a function that considers both
the status of a node set in relation to the larger graph and the
status of the individual nodes in a set in relation to one another.
The node set extraction is done by a greedy algorithm, and can deliver
different node sets according to the relative importance attached to
the scoring criteria, as well as to their detailed definition.

   The generation step synthesises the output summary text from the
selected predications. This may, however, be text only in a rather
restricted sense, since the data for doing the synthesis is limited.
This is because the individual source predications may be only fragments,
whether because source parsing has failed, or because the simple
predications derived from the sentence decomposition are incomplete:
in some cases atomic predications are more organisational devices than
substantial, and in some cases they are partial because referential
expressions are not resolved. Thus even
assuming that the method of identifying
important content has worked correctly, the final output summary is
best described as a `semi-text' indicative summary noting the main
topics of the source document. Clearly some method for organising and
ordering the output material is needed, but in the experimental work
so far with short sources and brief summaries this issue has not been
seriously addressed.
Thus the current procedure essentially groups predications with shared
arguments, constructs logical forms for these predication
clusters and any unattached predications, and applies the CLE to
synthesise output. This may consist of sentences or, for fragmentary
predications, mere phrases; and the presentation can
follow, as far as possible, the original source ordering of
the material.

   As a whole, the approach to summarising just described is a new
version of an old generic idea, i.e. that of representing a source text
as a network and looking for key nodes or node structures
(see e.g. Skorohod'ko 1972; Benbrahim and Ahmad 1994),
which has also now been explored as a means of summarising
multiple documents (Salton et al. 1997). But this network
view has usually been applied at the text level, with links
between surface or slightly normalised words or expressions, though
there may have been an assumption, especially with the latter,
that these were or were approximations to
predications.
Taylor (see Taylor and Krulee 1977) envisaged using a
deeper source text analysis along with graph abstraction from the
resulting semantic network, but his approach seems to have been only
partially implemented and rather modestly tested.
The major step taken in the research just described
has been to be much more thorough about the source text interpretation
into logical form, and to carry through a computational implementation
(if only a laboratory trial one) using modern NLP techniques,
as well as to explore the graph-processing possibilities more fully.

   Tucker's strategy clearly depends on some assumptions
and raises some issues; and both this specific approach to shallow
summarising using intermediate source representations and graph
structures, and others in the same broad class, have in particular
to be evaluated by comparison with surface text extraction methods,
whether these are ones delivering whole sentences or just providing
simple key term lists.

   The key assumptions are 
\begin{enumerate}
\item
  that source text content is intrinsically complex, i.e. involves
substantive meaning relations between concepts;
\item
  that (for many purposes) a summary that `covers' what is important
for a source as a whole is required; and
\item
  that in order to convey structured content, a summary also has to
be a text.
\end{enumerate}

   These assumptions may not seem controversial, but at least some work on
summarising rejects them. They imply that summarising
requires text meaning representations; and the argument is
that even though general-purpose, purely linguistic text
processing can only deliver meaning representations that are weak in
principle (as well as defective in practice), these are still superior to
source texts themselves for summarising precisely because they
support the identification of local and global topic structure,
which is grounded in
predicate-argument relational units.
The presumption is that
topics have structure that has to be captured and conveyed in a
more sophisticated and explicit way than by simple word
occurrences and coocurrences; this begins with individual elementary
predication units and extends to connected predications.
Further, though the basic predications are underspecified, and
the graph links between nodes that assume predicate
or argument similarity do this only in a non-referential way,
this is sufficient to {\em indicate} topics.
The graph operations naturally aggregate information,
which is taken further
in synthesis by clustering and a preference for more inclusive
predication structures; there is also some generalisation, in the
sense that selection omits detail. 
Finally, the way logical form decomposes text into multiple predications
supplies a better frequency marking for important entities, to anchor
the whole process of identifying summary material, than simple
word frequency.

   Results from the work are currently being evaluated. In the absence
of a task context, evaluation is limited to simple acceptability
to human readers, and comparisons with statistically-based extraction
methods. Thus the system can be used as a way of identifying key sentences,
i.e. those manifesting key predicate-argument structures, which can
be compared with the output from  simpler, purely statistical techniques.
Again, since its ordinary output is rather minimal as text, this can
be compared, if only informally, with phrases obtained by proximity or
very shallow parsing methods. (Whether this type of output is more useful
than extracted paragraph-length passages, or than visualisations of
underlying graph structures, requires a functional context.)
In the work described, there are many system parameters, e.g. the
link weighting scheme, the definition of representative subgraph,
so many comparative experiments within its own framework are required,
as well as tests against outputs from other general-purpose approaches.

   But more broadly, to test the general claims that underlie
this whole approach, there are many important problems and issues
to address. These include
\begin{enumerate}
\item
   coping with analysis limitations (e.g. minimal word sense
disambiguation) and failures (e.g. fragmentary representations);
\item
   incorporating direct content significance markers like
cue expressions, and indirect ones like discourse structure
indicators;
\item
   exploiting direct information about lexical frequencies,
i.e. statistical data;
\item
   taking advantage of additional general-purpose resources, e.g.
thesauri, term hierarchies;
\item
   extracting large-scale topic organisation from the basic
network;
\item
   addressing output presentation mechanisms, whether referring to
source constraints or new summary ones;
\item
   determining tradeoffs between computational effort and output
utility.
\end{enumerate}

   It is not evident precisely what may be useful or attainable:
for instance the analogy with document indexing suggests that
the lack of sense disambiguation may matter less than is assumed,
because lexical conjunction achieves it sufficiently for the user,
even if word meanings remain broad. But these questions have
to be explored. In addition, there is a need for fully functional
evaluation.

\section{Conclusion}

   This research programme is feasible, and it will be worthwhile if
it leads to better (i.e. more useful) summaries than extraction-based
methods, for which Brandow, Mitze and Rau (1995) may be taken as
representative.
Thus my argument is that pushing forward with shallow
summarising strategies
of the kind described has three important advantages. We already have
the NLP technology to start; we should get something that is
practically valuable; and we can learn from the tougher cases. Thus we
may gain insight into discourse structure through using predication
networks, and we may be able to gain insight into the best ways
of exploiting what in practice may be very limited domain information.

   Finally, the approach has the merit of being naturally extensible
or adaptable to longer texts, and in particular longer texts than ones
often encountered in summarising aimed at fair source coverage.
The claim on which
a shallow approach like that described is based is that what is
important in a text will `shine through' rough and partial source
interpretation. Longer text summarisation has to be addressed, because
the need for summaries is stronger, while the challenge for reduction
is greater. In the type of approach adopted, statistics and and
markedness clues can both be expected to be more in evidence and hence
more readily exploitable. Thus there should be a natural route forward
for scaleable summarising.

\end{document}